\documentclass[default,iicol]{sn-jnl}

\jyear{2021}%

\theoremstyle{thmstyletwo}%

\theoremstyle{thmstylethree}%

\usepackage{graphicx}
\usepackage{epsfig}
\usepackage{epstopdf}
\usepackage{stackengine,color}
\usepackage{amsmath,amssymb}
\begin{document}

\title[Symmetry Detection]{Robust Extrinsic Symmetry Estimation in 3D Point Clouds}


\author*[1]{\fnm{Rajendra} \sur{Nagar}}\email{rn@iitj.ac.in}

\affil*[1]{\orgdiv{Department of Electrical Engineering}, \orgname{Indian Institute of Technology Jodhpur}, \orgaddress{ \city{Jodhpur}, \postcode{342037}, \state{Rajasthan}, \country{India, rn@iitj.ac.in}}}

%


\abstract{Detecting the reflection symmetry plane of an object represented by a 3D point cloud is a fundamental problem in 3D computer vision and geometry processing due to its various applications, such as compression, object detection, robotic grasping, 3D surface reconstruction, etc. There exist several efficient approaches for solving this problem for clean 3D point clouds. However, it is a challenging problem to solve in the presence of outliers and missing parts. The existing methods try to overcome this challenge mostly by voting-based techniques but do not work efficiently. In this work, we proposed a statistical estimator-based approach for the plane of reflection symmetry that is robust to outliers and missing parts. We pose the problem of finding the optimal estimator for the reflection symmetry as an optimization problem on a 2-Sphere that quickly converges to the global solution for an approximate initialization. We further adapt the heat kernel signature for symmetry invariant matching of mirror symmetric points. This approach helps us to decouple the chicken-and-egg problem of finding the optimal symmetry plane and correspondences between the reflective symmetric points. The proposed approach achieves comparable mean ground-truth error and 4.5\% increment in the F-score as compared to the state-of-the-art approaches on the benchmark dataset.}




\maketitle

\section{Introduction}\label{sec1}
Real-world objects exhibit various kinds of symmetry. For example, architectural sites and houses exhibit reflection and translation symmetry, and human-made objects, such as vehicles, household objects, and furniture, exhibit rotational and reflection symmetry. Apart from human-made objects, most living objects, such as humans, animals, and insects, also exhibit reflection symmetry. The role of symmetry present in objects is to make objects physically balanced for smooth navigation and make objects visually attractive.   
\begin{figure}[!h]
	\centering
	\stackunder{\epsfig{figure=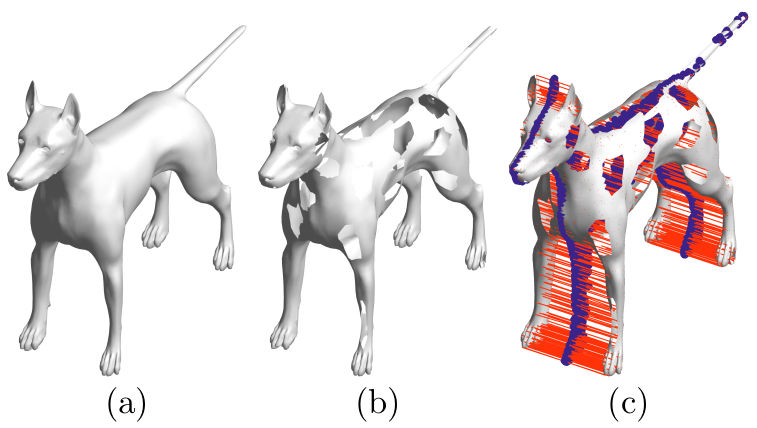,width=1\linewidth}}{}
	\caption{(a) Reflection symmetry of complete objects can be found efficiently \cite{funk20172017}. However, the 3D point cloud of an object can be partial due occlusions as shown in (b). Detecting symmetries from partial point clouds is a challenging and an open problem. In this work, we propose a statistical estimation-based  algorithm for efficiently detecting symmetry of partial objects as shown in (c).}
	\label{fig:intro}
\end{figure}

The fundamental problems in computer vision, computer graphics, and geometry processing mainly focus on processing and analyzing objects such as object detection and classification, physics-based photo-realistic rendering of objects, and 3D shape matching and texture transfer. Efficiently solving these problems requires a feature-level understanding of objects. The symmetry of objects plays a critical role in solving these problems \cite{mitra2013symmetry,liu2010computational,tyler2003human,zabrodsky1995symmetry} and many other application problems such as reconstructing 3D model of face from a single image \cite{wu2020unsupervised}, estimating 3D pose \cite{hodan2020epos}, image matching and recognition \cite{hauagge2012image}, face expression classification \cite{mitra2004local}, reconstructing 3D models of objects from a single image \cite{koser2011dense,miglani2013symmetry,thrun2005shape}, spectral shape correspondences and texture transfer \cite{melzi2019zoomout,ren2020maptree}, 3D registration \cite{hu2021globally}, 3D Shape Completion \cite{sung2015data,abbasi2019deep} and image re-colorization \cite{Lukac17-SIG}. 

Symmetry is classified into three fundamental categories: reflection, rotation, and translation. Also, depending on the amount of occlusion and noise present while acquiring a digital representation of an object, we can classify the symmetry as approximate or exact symmetry, partial or complete symmetry, and single or multiple symmetries. Also, the symmetry can be classified further as extrinsic symmetry (rigid objects such as a building) or intrinsic symmetry (dynamic objects such as animals and humans) based on the distance preserved (Euclidean or geodesic distance, respectively) under self-isometry transformation. An object can be represented in many formats, such as digital images, 3D point clouds, triangle meshes, implicit surfaces, etc. The problem of detecting various types of symmetries in different forms of object representations has been an active problem of research in the vision and graphics communities \cite{mitra2013symmetry,liu2010computational,funk20172017}. 

In this work, we propose an algorithm to find the global reflection symmetry plane of an object represented as a 3D point cloud acquired through a 3D scanner. We solve this problem in the presence of noisy points and missing parts of the objects. In literature, many efficient algorithms exist for detecting exact and complete symmetry of objects represented as 3D point clouds \cite{funk20172017}. However, detecting approximate and partial symmetry is an open problem \cite{funk20172017} and lacks a generalized formulation for detecting this form of symmetry. Most of the existing methods follow voting-based strategies for handling outliers and missing parts but fail to detect symmetry in the presence of a large number of outlier points and large missing parts \cite{mitra2006partial,lipman2010symmetry,shi2016symmetry}. Also, these methods construct a symmetry affinity matrix between the points that become intractable for large point clouds. 

The state-of-the-art approaches \cite{nagar2019detecting,ecins2017detecting,cicconet2017finding} pose the problem of detecting the plane of symmetry as an optimization problem. They use the L2-norm as a metric to find the residuals and assume that the set of correspondences between the reflective symmetric points does not contain many outliers correspondences. However, in practice, the set of correspondences between the mirror reflective points may contain many outliers correspondences due to sensor noise, missing parts, and non-ideal feature descriptors. Also, the approaches proposed in \cite{ecins2017detecting,nagar2019detecting,cicconet2017finding} used iterative closest point (ICP) based approaches where the plane of symmetry and correspondences between symmetric points are found simultaneously using an iterative algorithm. Since the L2-norm is not robust to outliers, the efficiency of these methods degrades in challenging settings. 
In this work, we propose a statistical estimator for the plane of reflection symmetry that is robust to outliers and missing parts efficiently. The idea is to use  $L_2E$ estimator proposed in \cite{basu1998robust} that has been used (\cite{tan2012automatic,jian2010robust,ma2013robust,mancas2005fast}) to solve many other problems in computer vision and computer graphics. The penalty curve
for the L2 norm is quadratic and assigns high probabilities even for outliers. Hence, its performance is affected by the outliers. Whereas the $L_2E$ estimator assigns very low probabilities for many of the residuals that help it be efficient in the presence of outliers. We further decouple the problem of finding the correspondences and the plane of symmetry to make the proposed approach computationally efficient. We first estimate a set of putative correspondences between mirror symmetric points. We then propose an approach to find reflection symmetry invariant descriptors for points. We use the heat diffusion characteristics (Heat Kernel Signature \cite{sun2009concise}) for finding descriptors for points. Since the heat diffusion process is an intrinsic property of the shape which is invariant to rigid self-isometry \cite{ovsjanikov2008global}, the proposed HKS-based descriptor, we name it Sym-HKS, is reflection symmetry invariant. Then, we pose the problem of finding an estimator for the plane of reflection symmetry and its distance from the origin as an optimization problem on a unit 2-Sphere. We show that the proposed approach achieves state-of-the-art performance on the benchmark dataset \cite{funk20172017}. In Figure \ref{fig:intro}, we show an example result. \\ \textbf{Contributions:} Our main contributions are the following.
\begin{enumerate}
	\item 	An HKS-based 3D point descriptor that is invariant to reflection symmetry transformation and a symmetry-aware matching method for finding mirror symmetric correspondences.  
	\item Statistical estimators for the reflection symmetry plane and its distance from the origin, which are robust to a significant number of outliers and missing parts due to occlusions.  
	\item We formulate the problem of finding the estimator as an optimization problem on 2-Sphere and then solve it using an optimization on manifolds technique.
\end{enumerate}
The remaining sections of the papers are organized as follows. In Section \ref{sec:RW}, we report the relevant research available in the literature and their limitations towards solving the symmetry detection problem efficiently. In Section \ref{sec:PA}, we describe the proposed algorithm for detecting reflection symmetry.  In Section \ref{subsec:rspe}, we define a manifold optimization problem for finding an estimator for the plane of reflection symmetry that is robust to outliers and missing parts. In Section, \ref{sec:SD}, we describe each point of the point cloud using a proposed descriptor that is reflection symmetry invariant descriptor. In Section \ref{RE}, we describe the evaluation procedure and the evaluation for measuring the performance of the proposed approach on the benchmark dataset.
\section{Related Work}  
\label{sec:RW} 
The extrinsic reflection symmetry detection problem in 3D objects represented by triangle meshes and  3D point clouds is a hot topic of research in the computer vision and graphics and 3D geometry processing community due to its various algorithmic advantages in terms of computational complexity, geometric representation, and applications \cite{funk2016symmetry,mitra2013symmetry}.  
In computer graphics and geometry processing literature, there exist many automatic algorithms for detecting the plane of reflection symmetry. 

The approach proposed in Zabrodsky \emph{et al.} \cite{zabrodsky1995symmetry} detect the reflection symmetry plane in a 2D point set. However, it needs a set of correct correspondences between mirror symmetric points computed in advance. The approaches proposed by  Lipman \emph{et al.}  \cite{lipman2010symmetry} and Xu \emph{et al.}  \cite{xu2012multi} find mirror-symmetric points using surface normals. They follow a voting-based approach and find symmetry orbits but need to tune various hyper-parameter. Also, for the approach proposed in \cite{lipman2010symmetry}, finding the symmetry factored embedding matrix becomes computationally intractable for larger size point clouds. Nagar and Raman used an optimization-based approach to find the plane of reflection symmetry in 3D point clouds with the provable guarantee of convergence of their method with a proper initialization \cite{nagar2019detecting}.  However, this approach requires solving an integer linear program that becomes computationally intractable for large point clouds and does not detect partial symmetry.  The method proposed in \cite{li2016efficient} uses multiple viewpoints of a 3D model to detect the plane of reflection symmetry. Whereas our approach can detect symmetry using only one viewpoint. The algorithm proposed by Comb{\'e}s \emph{et al.} \cite{combes2008automatic} detects reflection symmetry in a given 3D point cloud with a mild presence of outliers. This approach requires many hyper-parameters to be tuned, and it is computationally intractable as it requires solving an expectation maximization based problem at multiple scales.

The symmetry detection algorithms proposed  by  Mitra et al. \cite{mitra2006partial}, Speciale et al. \cite{speciale2016symmetry}, and Shi et al. \cite{shi2016symmetry}  use  point features to find reflection symmetry on 3D point clouds. These approaches use a voting-based approach to find the reflection symmetry plane that depends on the proper parameterization of the transformed space and require proper detection of modes in the transformed domain. Also, these approaches require a large number of pairs of points for voting that become intractable for large point clouds. Method proposed by Martinet \emph{et al.} \cite{martinet2006accurate} uses moment functions to detect symmetry and the method proposed by Berner \emph{et al.} \cite{berner2008graph} finds reflection symmetry using a graph constructed based on slippage features. However, these require graph connectivity for the input point clouds. The method proposed by Cohen \emph{et al.} \cite{cohen2012discovering}  uses the image features of pixels to find reflection symmetry in the structure from motion framework. However, it depends on image features to find symmetry efficiently. Cicconet  \emph{et al.}  \cite{cicconet2017finding} proposed a 3D registration based-approach. They first reflect the input point cloud about a random symmetry plane and then register both the point clouds to find the reflection symmetry plane. Ecins \emph{et al.} proposed a symmetric model fitting-based algorithm that is not robust to missing parts \cite{ecins2017detecting}. Also, they require segmentation of the symmetric object to find the symmetry. These two algorithms  \cite{cicconet2017finding} and \cite{ecins2017detecting} formulate  the problem of  symmetry estimation as a 3D rigid registration problem that increases the parameters of the reflection symmetry transformation as rotation matrix has three parameters for 3D registration, but the reflection symmetry plane has only two parameters to estimate. This leads an increased computational complexity. Hruda \emph{et al.} proposed an efficient method where they proposed a differential measure for reflection symmetry in 3D point clouds \cite{hruda2020plane,hruda2021robust}. However, the performance of this method does not perform well for non-uniformly sampled point clouds. Also, it is robust to missing parts but may fail in presence of many scattered outlier points. Whereas, the proposed approach is robust to such cases.

There are various exciting approaches that use surface features to detect symmetry, such as \cite{lasowski2009probabilistic,thomas2014multiscale,kazhdan2002reflective}. However, these approaches can not directly be adapted to work on volumetric point clouds as the features used in these approaches assume that the point cloud is sampled from a 2-manifold surface.  Ovsjanikov \emph{et al.} \cite{ovsjanikov2008global}, Qiao \emph{et al.} \cite{qiao2019learning} Wang \emph{et al.} \cite{wang2017group}, Nagar and Raman \cite{nagar2018fast}, Xu \emph{et al.} \cite{xu2009partial}, Wang \emph{et al.} \cite{wang2019intrinsic}, Liu \emph{et al.} \cite{liu2012finding}, Sahilliouglu \emph{et al.} \cite{sahilliouglu2013coarse}, and Sipiran \emph{et al.} \cite{sipiran2014approximate} use spectral properties of the  Laplace-Beltrami operator  and Kim \emph{et al.} \cite{kim2010mobius} use M\"{o}bius transformation to detect intrinsic symmetries of 3D objects represented by triangle meshes. However, these methods do not generalize to find extrinsic symmetry in 3D point clouds as they depend on mesh connectivity between points and assume that the underlying object is represented by a 2-manifold surface.  The problem of reflection symmetry detection in digital images also is an active field of research \cite{funk20172017,gnutti2021combining,seo2021learning,zhou2021nerd,shi2020symmetrynet}. However, these methods may not be generalized to detect 3D symmetry in point clouds due to completely different representations of objects. The problem of symmetry plane estimation and correspondences estimation is chicken-and-egg problem. In order to solve this, we first have to find a set of  correspondences between mirror symmetric points. Estimating correspondences between symmetric points is a challenging problem to solve \cite{van2011survey, sahilliouglu2020recent}.  

Recently, the problem of detecting reflection symmetry in 2D and 3D data has been addressed using learning based techniques. Shi \emph{et al.} detect reflection as well as rotational symmetry in RGB-D data \cite{shi2020symmetrynet}. Zhou \emph{et al.} learn a neural detector for detecting 3D symmetry of an object from a single RGB image \cite{zhou2021nerd}. Seo \emph{et al.} use polar matching convolution for detecting 2D symmetries in an RGB image \cite{seo2021learning}. The current state-of-the-art method by Gao \emph{et al.} uses neural networks for detecting 3D symmetries in point cloud data \cite{gao2020prs}. 
\section{Proposed Approach}
\label{sec:PA}
\subsection{Problem Formulation}
\label{subsec:pf}
Let $\mathcal{P}=\{\mathbf{x}_i\}_{i=1}^n\subset \mathbb{R}^3$ be a noisy and incomplete point cloud sampled from the surface of a symmetric object. Let $\mathbf{v}\in\mathbb{R}^3$ be a unit norm  vector that is normal to the plane of reflection symmetry and $\omega$ be the distance of the symmetry plane from the origin. We can mathematically model the reflection symmetry using  the Householder transform. That is, if $\mathbf{x}$ and $\mathbf{y}$ are reflective symmetric points, then $\mathbf{y}=(\mathbf{I}-2\mathbf{vv}^\top)\mathbf{x}+2\omega\mathbf{v}$. Given the point cloud $\mathcal{P}$, we  pose the problem of estimating $\omega$ and $\mathbf{v}$ as the below optimization problem.
\begin{equation}
	\underset{\underset{ \mathbf{v}^\top\mathbf{v}=1}{\omega\in\mathbb{R},\mathbf{v}\in\mathbb{R}^3}}{\arg\min}\sum_{i=1}^{n}\|\mathbf{x}_{\pi(i)}-(\mathbf{I}-2\mathbf{vv}^\top)\mathbf{x}_i-2\omega\mathbf{v}\|_2^2.
	\label{eq1}
\end{equation}
Here, $\mathbf{x}_{\pi(i)}$ denotes the reflection of the point $\mathbf{x}_i$ about the  symmetry plane. We can easily find the optimal $\omega$ that minimizes this optimization problem by setting the derivative of the cost function which is defined as $\omega=\frac{1}{n}\sum_{i=1}^n\mathbf{v}^\top(\frac{\mathbf{x}_i+\mathbf{x}_{\pi(i)}}{2})$. Now, we replace this optimal $\omega$ back in the optimization function and  we rewrite the above formulation by observing that  $\|\mathbf{x}_{\pi(i)}-(\mathbf{I}-2\mathbf{vv}^\top)\mathbf{x}_{i}-2\omega\mathbf{v}\|_2^2=4\mathbf{v}^\top\mathbf{x}_i\mathbf{x}_{\pi(i)}^\top\mathbf{v}+\|\mathbf{x}_i-\mathbf{x}_{\pi(i)}\|_2^2-4\omega\mathbf{v}^\top(\mathbf{x}_{\pi(i)}+\mathbf{x}_i)+4\omega^2$.  Therefore, the optimization problem  defined in Equation \eqref{eq1}  for solving for $\mathbf{v}$ is equivalent to the problem defined in Equation \eqref{eq2}. 
\begin{equation}
	\mathbf{v}^\star=\underset{\mathbf{v}\in\mathbb{R}^3, \mathbf{v}^\top\mathbf{v}=1}{\arg\min}\mathbf{v}^\top\mathbf{H}\mathbf{v}.
	\label{eq2}
\end{equation}
Here, the matrix $\mathbf{H}\in\mathbb{R}^{3\times 3}$ is defined as $\mathbf{H}=\sum\limits_{i=1}^{n}\mathbf{x}_i\mathbf{x}_{\pi(i)}^\top$. A closed form solution to this problem can easily be found by minimizing the function $\mathbf{v}^\top\mathbf{H}\mathbf{v}+\lambda(\mathbf{v^\top v}-1)$. The optimal vector $\mathbf{v}^\star$ that minimizes this function is the eigenvector of the matrix $\mathbf{H}$ corresponding to the smallest eigenvalue. In order to find the matrix $\mathbf{H}$, we should know the reflection point $\mathbf{x}_{\pi(i)}$ for each point $\mathbf{x}_i$. However, we can not find the point of reflection without knowing the plane of reflection symmetry. Therefore,  we have to solve two coupled problems: Find the symmetry plane $(\mathbf{v},\omega)$  using the mirror symmetric correspondences $\{(\mathbf{x}_i,\mathbf{x}_{\pi(i)})\}_{i=1}^n$ and find the mirror symmetric correspondences $\{(\mathbf{x}_i,\mathbf{x}_{\pi(i)})\}_{i=1}^n$ using the plane of reflection symmetry $(\mathbf{v},\omega)$. Here, both $(\mathbf{v},\omega)$ and $\{(\mathbf{x}_i,\mathbf{x}_{\pi(i)})\}_{i=1}^n$ are unknown. Hence, the problem defined in Equation \eqref{eq2} requires both correspondences as well as the plane of reflection symmetry which depend on each other. We know that given the correspondences between the mirror symmetric points, finding the optimal $(\mathbf{v},\omega)$ is easy as it is the eigenvector corresponding to the smallest eigenvalue of the matrix $\mathbf{H}$. However, finding correspondences $(\mathbf{x}_i,\mathbf{x}_{\pi(i)})$ amounts to solving a linear assignment problem. This can become intractable for large point clouds. In order to solve this problem efficiently, we make use of surface features. In Section \ref{sec:SD}, we propose an approach to find symmetry aware feature descriptors for the input point cloud. 
\subsection{Robust Symmetry Plane Estimation}
\label{subsec:rspe}
The set of putative correspondences we get through matching the reflection invariant feature descriptors may contains many outlier correspondences. The reason for this is that there can be many outlier points present in the given point cloud. Also, a some parts of the object may be missing, that leads to wrong matches for points whose mirror reflective points are missing. The optimal symmetry plane $(\mathbf{v},\omega)$, found by solving the optimization problem defined in Equation \eqref{eq2}, as  L2-norm is not robust to outliers.\\
In order to find the optimal symmetry plane $(\mathbf{v},\omega)$, we propose a statistical estimation technique based on the  $L_2E$ estimator. Let $\{(\mathbf{x}_i,\mathbf{x}_{\pi(i)})\}_{i=1}^n$ be a set of correspondences between mirror-symmetric points that may contain outliers correspondences. Now, we model the estimation problem as follows:
\begin{equation}
	\mathbf{x}_{\pi(i)}=(\mathbf{I-2\mathbf{v}\mathbf{v}^\top})\mathbf{x}_i+2\omega\mathbf{v}+\boldsymbol{\epsilon},\; i\in\{1,\ldots,n\}.
	\label{eq3_0}
\end{equation}

Here, $\boldsymbol{\epsilon}\in\mathbb{R}^3$ is noise and we assume that it follows the Gaussian distribution with zero mean and $\sigma^2$ variance. I.e., $\boldsymbol{\epsilon}\sim\mathcal{N}(\mathbf{0},\text{diag}(\begin{bmatrix}\sigma^2&\sigma^2&\sigma^2\end{bmatrix}))$ and all its samples are identically and independently distributed. Let us assume that the true  symmetry plane parameters are $(\mathbf{v}_0,\omega_0)$ $\mathbf{v}_0$. Now, let $g(\mathbf{r}\mid\mathbf{v},\omega)$ be the parametric density model with respect to the estimation parameters $\mathbf{v}$ and $\omega$. Let $g(\mathbf{r}\mid\mathbf{v}_0,\omega_0)$ be the parametric density model with respect to the true parameters $\mathbf{v}_0$ and $\omega_0$. Here, $\mathbf{r}$ denotes the residual vector. Then, the $L_2E$ estimator $\hat{\mathbf{v}}$ for $\mathbf{v}_0$ and $\hat{\omega}$ for $\omega$ are found by minimizing the loss $\int(g(\mathbf{r}\mid\mathbf{v},\omega)-g(\mathbf{r}\mid\mathbf{v}_0,\omega))^2d\mathbf{r}$ with respect to $\mathbf{v}$ and $\omega$. Following the theory of L2E \cite{basu1998robust}, the optimal $\hat{\mathbf{v}}$ and $\hat{\omega}$  can be found by solving the optimization problem defined in Equation \eqref{eq4} with respect to $\mathbf{v}$ and $\omega$.
\begin{align}
	&\underset{\underset{\mathbf{v^\top v}=1}{\omega\in\mathbb{R},\mathbf{v}\in\mathbb{R}^3}}{\min}\int g^2(\mathbf{r}\mid\mathbf{v},\omega)d\mathbf{r}-\frac{2}{n}\sum_{i=1}^ng(\mathbf{r}_i\mid\mathbf{v},\omega).
	\label{eq4}
\end{align}
Since we assume that the $\boldsymbol{\epsilon}$ is the white Gaussian noise the residual vectors $\mathbf{r}_i=\mathbf{x}_{\pi(i)}-(\mathbf{I}-2\mathbf{vv}^\top)\mathbf{x}_i-2\omega\mathbf{v}$ follow a Gaussian distribution, i.e., $g(\mathbf{r}_i\mid\mathbf{v},\omega)=\frac{1}{(2\pi\sigma^2)^{\frac{3}{2}}}e^{-\frac{\|\mathbf{x}_{\pi(i)}-(\mathbf{I}-2\mathbf{vv}^\top)\mathbf{x}_i-2\omega\mathbf{v}\|_2^2}{2\sigma^2}}$. Also, since $g(\mathbf{r}\mid\mathbf{v},\omega)$ is a Gaussian distribution the term  $\int g^2(\mathbf{r}\mid	\mathbf{v},\omega)d\mathbf{r}=\frac{1}{(2\pi\sigma^2)^{\frac{3}{2}}}$ which is constant with respect to the parameters $\mathbf{v}$ and $\omega$. Hence, the $L_2E$ estimators $\hat{\mathbf{v}}$ and $\hat{\omega}$  for $\mathbf{v}$ and $\omega$, respectively, can be defined as in Equation \eqref{eq6}. 
\begin{align}
	\nonumber&=\underset{\underset{\mathbf{v^\top v}=1}{\omega\in\mathbb{R},\mathbf{v}\in\mathbb{R}^3}}{\max}\sum_{i=1}^n\frac{2}{n(2\pi\sigma^2)^{\frac{3}{2}}}e^{-\frac{\|\mathbf{x}_{\pi(i)}-(\mathbf{I}-2\mathbf{vv}^\top)\mathbf{x}_i-2\omega\mathbf{v}\|_2^2}{2\sigma^2}}\\
	&=\underset{\underset{\mathbf{v^\top v}=1}{\omega\in\mathbb{R},\mathbf{v}\in\mathbb{R}^3}}{\max}f(\mathbf{v},\omega).
	\label{eq6}
\end{align}

Now, in order to find the optimal estimators $(\hat{\mathbf{v}},\hat{\omega})$  for $(\mathbf{v},\omega)$, we have to find the maximum of the function $f$ with respect to $\mathbf{v}$ and $\omega$. We follow an \textit{alternating optimization} approach to find the optimal parameters. We first initialize $\mathbf{v}$ (which we describe at the end of this section) and find optimal $\omega$. Then, we update $\mathbf{v}$ with the new $\omega$. We keep alternating between these two optimization problems till convergence. We first describe the approach for finding optimal $\mathbf{v}$ given $\omega$.\\
\textbf{Optimal $\mathbf{v}$}: We observe that the domain of the function $f$ is the 2-Sphere $\mathbb{S}^{2}=\{\mathbf{v}\in\mathbb{R}^3\mid\mathbf{v}^\top\mathbf{v}=1\}$, which is a smooth 2-manifold \cite{absil2009optimization}. Therefore, in order to solve the problem defined in Equation \eqref{eq6} with respect to $\mathbf{v}$, we use the technique of optimization on manifolds  \cite{absil2009optimization}. We find the Riemannian gradient of the function $f$ on the manifold $\mathbb{S}^2$  by first finding the Euclidean gradient on the tangent plane and then projecting it to the sphere as follows. Let $\nabla f(\mathbf{v})$ be the Euclidean gradients and $\text{grad} f(\mathbf{v})$ be the Riemannian gradient of the function $f$. Then, the Riemannian  and the Euclidean gradients are related as $\text{grad} f(\mathbf{v})=\mathbb{P}_\mathbf{v}(\nabla f(\mathbf{v}))$. Here, $\mathbb{P}_\mathbf{v}$ is the projection operator for $\mathbb{S}^2$ that projects an ambient space vector onto the  tangent space at $\mathbf{v}$ and defined as $\mathbb{P}_\mathbf{v}(\mathbf{x})=\mathbf{x}-(\mathbf{x}^\top\mathbf{v})\mathbf{v}$. 	The Euclidean gradient of the function $f(\mathbf{v},\omega)=\frac{2}{n(2\pi\sigma^2)^{\frac{3}{2}}}\sum\limits_{i=1}^ne^{-\frac{f_i(\mathbf{v},\omega)}{2\sigma^2}}$ can easily be found by and is defined in Equation \eqref{eqq6}. Here, $f_i(\mathbf{v},\omega)=4\mathbf{v}^\top\mathbf{C}_i\mathbf{v}+\|\mathbf{x}_i-\mathbf{x}_{\pi(i)}\|_2^2-4\omega\mathbf{v}^\top(\mathbf{x}_{\pi(i)}+\mathbf{x}_i)+4\omega^2$. 

\begin{align}
	\nonumber \nabla f(\mathbf{v})&=\sum\limits_{i=1}^n\frac{8\omega\mathbf{m}_i-4(\mathbf{C}_i+\mathbf{C}_i^\top)\mathbf{v}}{n\sigma^2(2\pi\sigma^2)^{\frac{3}{2}}}e^{-\frac{f_i(\mathbf{v},\omega)}{2\sigma^2}}\\
	&=\frac{4(\mathbf{1}_n^\top\otimes\mathbf{I}_3)[( \mathbf{A}\odot(\omega\mathbf{D}-\mathbf{B}(\mathbf{I}\otimes\mathbf{v}))]\mathbf{1}_n}{n\sigma^2(2\pi\sigma^2)^\frac{3}{2}}.
	\label{eqq6}
\end{align} 
Here, $\mathbf{m}_i=\frac{\mathbf{x}_{\pi(i)}+\mathbf{x}_i}{2}$, $\mathbf{C}_i=\mathbf{x}_i\mathbf{x}_{\pi(i)}^\top$, $\mathbf{A}\in\mathbb{R}^{3n\times n}$ is a block diagonal matrix where the $i$-th diagonal block is defined as $e^{-\frac{f_i(\mathbf{v},\omega)}{2\sigma^2}}\mathbf{1}_3$, the matrix  $\mathbf{B}\in\mathbb{R}^{3n\times 3n}$ is also a diagonal matrix where the $i$-th diagonal block is equal to $\mathbf{C}_i+\mathbf{C}_i^\top$,  the matrix  $\mathbf{D}\in\mathbb{R}^{3n\times n}$ is  a diagonal matrix where the $i$-th diagonal block is equal to $\mathbf{x}_i+\mathbf{x}_{\pi(i)}$, $\mathbf{I}$ is the identity matrix of size $n\times n$, $\mathbf{1}_n\in\{1\}^n$, $\otimes$ is the Kronecker matrix product operator, and $\odot$ is the element-wise matrix multiplication operator.  We use the Manifold-BFGS optimization algorithm, proposed in \cite{absil2009optimization},  for finding the optimal solution. We use the \textit{manopt} toolbox for optimization on manifolds \cite{boumal2014manopt}.\\ We further observe that the function $f$ is not a convex function and is locally convex around the global maximum. Therefore, the final solution and the convergence rate  depend on the initialization of vector $\mathbf{v}$. \\
\textbf{Optimal $\omega$}: Given the initialized $\mathbf{v}$, the function $f(\mathbf{v},\omega)$ is a non-convex function with respect to $\omega$. Therefore, we solve this non-linear and non-convex optimization problem using the BFGS (Broyden–Fletcher–Goldfarb–Shanno) Quasi-Newton algorithm \cite{fletcher2000practical}. We provide the exact and explicit derivative of the cost function $\frac{\partial f(\mathbf{v},\omega)}{\partial\omega}$ to make the BFGS optimization algorithm to converge faster, which is defined in Equation \eqref{eqnn}.
\begin{align}
	\frac{\partial f(\mathbf{v},\omega)}{\partial\omega}&=\frac{4}{n}\sum\limits_{i=1}^n\frac{\mathbf{v}^\top(\mathbf{x}_{\pi(i)}+\mathbf{x}_i)-2\omega}{\sigma^2(2\pi\sigma^2)^{\frac{3}{2}}}e^{-\frac{f_i(\mathbf{v},\omega)}{2\sigma^2}}.
	\label{eqnn}
\end{align} 
\textbf{Initialization of $\mathbf{v}$:} The normal vector to the symmetry plane is perpendicular to the plane containing the  mid-points of the to the vectors joining mirror symmetric points 
(i.e.,  $(\mathbf{x}_i)$ and $\mathbf{x}_{\pi(i)}$). Therefore, in case of clean and complete point clouds, we may initialize the vector $\mathbf{v}$ as the eigenvector of the matrix $\frac{1}{n}\sum\limits_{i=1}^n\left(\frac{\mathbf{x}_i+\mathbf{x}_{\pi(i)}}{2}\right)\left(\frac{\mathbf{x}_i+\mathbf{x}_{\pi(i)}}{2}\right)^\top$ corresponding to the smallest eigenvalue. This is the classical principle component analysis approach. However, in case of noisy correspondences, the classical PCA may fail to find the initial $\mathbf{v}$ approximately. We, therefore, use the  Grassmann manifold averaging based robust PCA algorithm \cite{hauberg2014grassmann}.  We use the Robust Grassmann Average (RGA) algorithm to find the vectors $\mathcal{\eta}_1$ and $\mathbf{\eta}_2$ that spans the subspace represented by the set of mid-points $\{\frac{\mathbf{x}_i+\mathbf{x}_{\pi(i)}}{2}\}_{i=1}^n$ of the line segments joining the mirror-symmetric points $\mathbf{x}_i$ and $\mathbf{x}_{\pi(i)}$. Then, we initialize $\mathbf{v}=\mathcal{\eta}_1\times\mathbf{\eta}_2$.\\
\textbf{A Remark on MLE}: We would like to clarify that the cost function, defined in Equation \eqref{eq6}, for the L2E estimator is different than the maximum likelihood estimator (MLE). We can easily determine the MLE estimator $\mathbf{v}_\mathrm{mle}$ as by solving the below optimization problem:
\begin{eqnarray}
	\nonumber\mathbf{v}_\mathrm{mle}&=&\underset{\mathbf{v}\in\mathbb{R}^3,\mathbf{v^\top v}\top=1 }{\arg\max}\frac{1}{(2\pi\sigma^2)^{\frac{3n}{2}}}e^{-\frac{2}{\sigma^2}\sum\limits_{i=1}^n \mathbf{v}^\top\mathbf{x}_i\mathbf{x}_{\pi(i)}^\top\mathbf{v}}\label{eq9}\\
	\mathbf{v}_\mathrm{mle}&=&\underset{\mathbf{v}\in\mathbb{R}^3,\mathbf{v^\top v}\top=1 }{\arg\min}\sum\limits_{i=1}^n \mathbf{v}^\top\mathbf{x}_i\mathbf{x}_{\pi(i)}^\top\mathbf{v}.
\end{eqnarray}
We can observe that the cost functions of $L_2E$ estimator and MLE estimators defined in Equation  \eqref{eq9}, respectively,  are different. In Section \ref{sec:caa}, we analyse  the robustness of the $L_2E$ and the MLE estimators in presence of outliers.  
\section{Mirror Symmetric Point Correspondences}
\label{sec:SD}

Our goal is to find a set of candidate matching between mirror symmetric points to find the plane of reflective symmetry plane. Since the proposed algorithm for finding the normal vector to the plane of reflection symmetry is robust to outlier matches, our aim is to find a set with a few correct correspondences where other correspondences can be outliers. In order to find a set $\{(\mathbf{x}_i,\mathbf{x}_{\pi(i)})\}_{i=1}^n$ of putative correspondences between the mirror-symmetric points, we need to find the mapping $\pi:\{1,2\ldots,n\}\rightarrow\{1,2,\ldots,n\}$. In order to find the matching $\pi$, we first find a set of feature points and their symmetry invariant descriptors and then solve a matching problem.\\
\textbf{Sym-HKS: Feature Point Detection and Description}. In order to find the mirror symmetric feature points, we use the Intrinsic Shape Signature (ISS) approach proposed by Zhong 2009 \cite{zhong2009intrinsic} as the feature points detected by this approach are intrinsic. Therefore, it detects intrinsically symmetric points on the given point cloud. Let $\{\mathbf{o}_i\}_{i=1}^a$ be the set of detected $a$ ISS keypoints. Now, we find a feature descriptor for each of the detected ISS keypoint using the heat kernel signatures technique Sun et al. 2009 \cite{sun2009concise}. There exist other approaches for feature description (such as \cite{li2021anisotropic,wang2019robust,wang2020mgcn}). However, we prefer HKS over the other approaches as it is reflection symmetry and scale invariant. The HKS requires the eigenfunctions of the Laplace Beltrami operator. Therefore, we first construct the Laplacian Beltrami operator of the surface defined by the input point cloud using the approach proposed by Sharp and Crane 2020 \cite{sharp2020laplacian}. There exist other approaches (such as \cite{belkin2009constructing},\cite{liu2012point}) for finding the LBO for point clouds, but they mostly work for manifold surfaces and not scale for non-manifold surfaces such as a chair. Most of the objects in practice have non-manifold surfaces. The Tufted Laplacian algorithm proposed in \cite{sharp2020laplacian} finds the Laplacian matrix and the vertex area matrix for the surface defined by the input points with requiring the points connectivity. We solve the generalized eigenvalue problem $\mathbf{L}\boldsymbol{\phi}_i=\lambda_i\mathbf{M}\boldsymbol{\phi}_i$ to find the eigenfunctions corresponding to the first $k$ smallest eigenvalues of the Laplace Beltrami operator. Here $\mathbf{L}$ is the sparse Laplacian matrix and $\mathbf{M}$ is the diagonal mass matrix. Now, the heat kernel signature $\mathbf{h}_i\in\mathbb{R}^p$ of the $i$-th ISS keypoint $\mathbf{o}_i$ is defined in Equation \eqref{eq99} 
\begin{align}
	\mathbf{h}_i&=\begin{bmatrix}\sum\limits_{j=1}^k e^{-t_0\lambda_j}\boldsymbol{\phi}^2_j(\mathbf{o}_i)&\cdots&\sum\limits_{j=1}^k e^{-t_{p-1}\lambda_j}\boldsymbol{\phi}^2_j(\mathbf{o}_i)\end{bmatrix}^\top.
	\label{eq99}
\end{align}
Here, $t_0,\ldots,t_{p-1}$ are $p$ time instances uniformly spaced in the range $\left[\frac{10\log_e(10)}{\lambda_k},\frac{10\log_e(10)}{\lambda_2}\right]$ as suggested in \cite{sun2009concise}.  \\
\textbf{Estimating Mirror Symmetric Correspondences}. In order to find a set of putative correspondences between mirror symmetric points, we can use the HKSs of the feature points as the HKS for mirror symmetric points would be the same. In order to find the mirror symmetric point of the keypoint $\mathbf{o}_i$, we can find the keypoint from the set $\{\mathbf{o}_i\}_{i=1}^a\backslash\{\mathbf{o}_i\}$ which has the closest HKS with that of the $\mathbf{o}_i$. However, this strategy may fail in general due to the following reason. Let us consider three points $\mathbf{o}_i$, $\mathbf{o}_i$, and $\mathbf{o}_{\ell}$ such that $\mathbf{o}_j$ is a neighbor of $\mathbf{o}_i$ and $\mathbf{o}_{\ell}$ is the mirror reflection of $\mathbf{o}_i$. Now, since $\mathbf{o}_i$ and $\mathbf{o}_j$ are neighbors and the eigenfunctions are at least twice differentiable (as they are the solution of the Laplace equation) hence $\boldsymbol{\phi}(\mathbf{o}_i)\approx\boldsymbol{\phi}(\mathbf{o}_j)$. Therefore,  the HKS  $\mathbf{h}_i$ and $\mathbf{h}_{j}$ would be similar as $\|\mathbf{h}_i-\mathbf{h}_{j}\|_2^2=\sum_{g=0}^{p-1}(\sum_{i=1}^ke^{-t_{g}\lambda_i}(\boldsymbol{\phi}_i^2(\mathbf{o}_i)-\boldsymbol{\phi}_i^2(\mathbf{o}_j)))^2\approx0$.\\ This may result in multiple neighbor matches $(\mathbf{o}_i, \mathbf{o}_i)$. Therefore, we must ensure that the nearby points are not matched. To enforce this constraint, we need a way of measuring the closeness of the points on the surface defined by the point cloud. Geodesic distance is one of the best options, but it would require connectivity of the points and may be computationally inefficient for large point clouds. \\We circumvent this challenge by using the eigenfunctions of the Laplace Beltrami operator as follows which we have already computed to find HKS. Let $\mathbf{s}_i=\begin{bmatrix}s_{i0}&s_{i1}&\cdots&s_{ia}\end{bmatrix}$ be the sign vector for the feature point $\mathbf{o}_i$ where $s_{ij}=+1$ if $\boldsymbol{\phi}_j(\mathbf{o}_i)\geq0$ and  $s_{ij}=-1$ if $\boldsymbol{\phi}_j(\mathbf{o}_i)<0$. Now, if the two points are neighbors, they would be in the same nodal set of low-frequency eigenfunctions. Therefore, $\mathbf{s}_i=\mathbf{s}_j$ for the two neighoring points $\mathbf{o}_i$ and $\mathbf{o}_j$. However, if we consider two mirror symmetric points $\mathbf{o}_i$ and $\mathbf{o}_{\ell}$ which lie on the two different halves of the objects, then they would lie on different nodal domains which might have different signs as some of the eigenfunctions of the Laplace Beltrami operator would be having negative sings (equivalent to odd functions in the Euclidean sense). Therefore, $\|\mathbf{s}_i-\mathbf{s}_{\ell}\|_2$ will be very high for the two mirror symmetric points. With this approach, we formulate the following optimization approach, which prohibits neighboring points matchings and ensures mirror symmetric points matching. 
\begin{eqnarray}
	\underset{\mathbf{P}\in\{0,1\}^{k\times k}}{\max} &&\sum_{i=1}^k\sum_{i=1}^k p_{ij}a_{ij}\\
	\text{subject to }\mathbf{1^\top P}&\leq&\mathbf{1}^\top\\
	\mathbf{P1}&\leq&\mathbf{1}\\
	\mathbf{1^\top P 1}&=&2q
\end{eqnarray}

Here, $a_{ij}=\|\mathbf{h}_i-\mathbf{h}_j\|_2+\psi(\|\mathbf{s}_i-\mathbf{s}_j\|_2)$, $\psi(t)=b$ if $t=0$ and $\phi(t)=0$ for $t>0$. Here, $b$ is a very large constant which we set equal to 1000 in our experiment.  The matrix  $\mathbf{P}\in\{0,1\}^{k\times k}$ is the binary matching matrix with its $(i,j)$-th entry defined as $p_{ij}=1$ if the feature points $\mathbf{o}_i$ and $\mathbf{o}_j$ form are mirror symmetric images of each other and $p_{ij}=0$, otherwise. The constant $q$ denotes the number of mirror symmetric correspondences we want, which would ensure that we choose the best $q$ candidate correspondences. This optimization problem is a standard linear integer program that we solve using the  \texttt{intlinprog} function in the MATLAB.
	\begin{algorithm}[!h]
	\caption{\textit{Robust Reflection Symmetry Plane Estimation}}\label{alg:0}
	\textbf{Input}: A 3D Point cloud $\mathcal{P}=\{\mathbf{x}_i\}_{i=1}^n$ 
	\begin{algorithmic}[1]
		\State $\mathbf{L}\gets\texttt{Tufted-LBO}(\mathcal{P})$ {\tiny{\color{blue}$\triangleright$ find LBO Eigenfunctions}}
		\State $\{(\mathbf{x}_i,\mathbf{x}_{\pi(i)})\}_{i=1}^n\gets\texttt{Sym-HKS}(\mathbf{L})$ {\tiny{\color{blue}$\triangleright$  correspondences}}
		\State $\mathbf{O}\gets\begin{bmatrix}\frac{\mathbf{x}_1+\mathbf{x}_{\pi(1)}}{2}&\cdots&\frac{\mathbf{x}_n+\mathbf{x}_{\pi(n)}}{2}\end{bmatrix}^\top$
		\State $\mathbf{v}\gets\texttt{RGA}(\mathbf{O})$  {\tiny{\color{blue}$\triangleright$  find initial $\mathbf{v}$}}
		\State  $\mathbf{C}_i\gets\mathbf{x}_{\pi(i)}\mathbf{x}_i^\top,\forall i\{1,2,\ldots,n\}$.
		\State $\mathbf{B}\gets\text{diag}\left(\begin{bmatrix}\mathbf{C}_1+\mathbf{C}_1^\top&\cdots&\mathbf{C}_n+\mathbf{C}_n^\top\end{bmatrix}\right)$
 \State $\mathbf{D}\gets\text{diag}\left(\begin{bmatrix}(\mathbf{x}_1+\mathbf{x}_{\pi(1)})&\cdots&(\mathbf{x}_n+\mathbf{x}_{\pi(n)})\end{bmatrix}\right)$
\While{not converged} {\tiny{\color{blue}$\triangleright$  Alternating optimization}}
\State $\omega\gets\texttt{BFGS}(f(\mathbf{v},\omega),\frac{\partial f(\mathbf{v}}{\partial \omega})$ {\tiny{\color{blue}$\triangleright$  find $\omega$ given $\mathbf{v}$}}
\While{not converged} {\tiny{\color{blue}$\triangleright$  find $\mathbf{v}$ given $\omega$}}
		\State $\mathbf{A}\gets\text{diag}\left(\begin{bmatrix}e^{-\frac{f_1(\mathbf{v},\omega)}{2\sigma^2}}\mathbf{1}_3&\cdots&e^{-\frac{f_n(\mathbf{v},\omega)}{2\sigma^2}}\mathbf{1}_3\end{bmatrix}\right)$
		\State $\boldsymbol{\beta}\gets\frac{4(\mathbf{1}_n^\top\otimes\mathbf{I}_3)[( \mathbf{A}\odot(\omega\mathbf{D}-\mathbf{B}(\mathbf{I}\otimes\mathbf{v}))]\mathbf{1}_n}{n\sigma^2(2\pi\sigma^2)^\frac{3}{2}}$
		\State $\text{grad} f(\mathbf{v})\gets\boldsymbol{\beta}-\boldsymbol{\beta}^\top\mathbf{v}\mathbf{v}$ {\tiny{\color{blue}$\triangleright$ Riemannian gradient}}	
		\State $\mathbf{v}\gets\texttt{Manifold-BFGS}(\text{grad} f(\mathbf{v}),f,\mathbf{v})$
		\EndWhile
\EndWhile
	\end{algorithmic}
	\textbf{Output:}  $\mathbf{v},\omega$.
\end{algorithm}
	\begin{figure*}[!h]
		\centering
\stackunder{\epsfig{figure=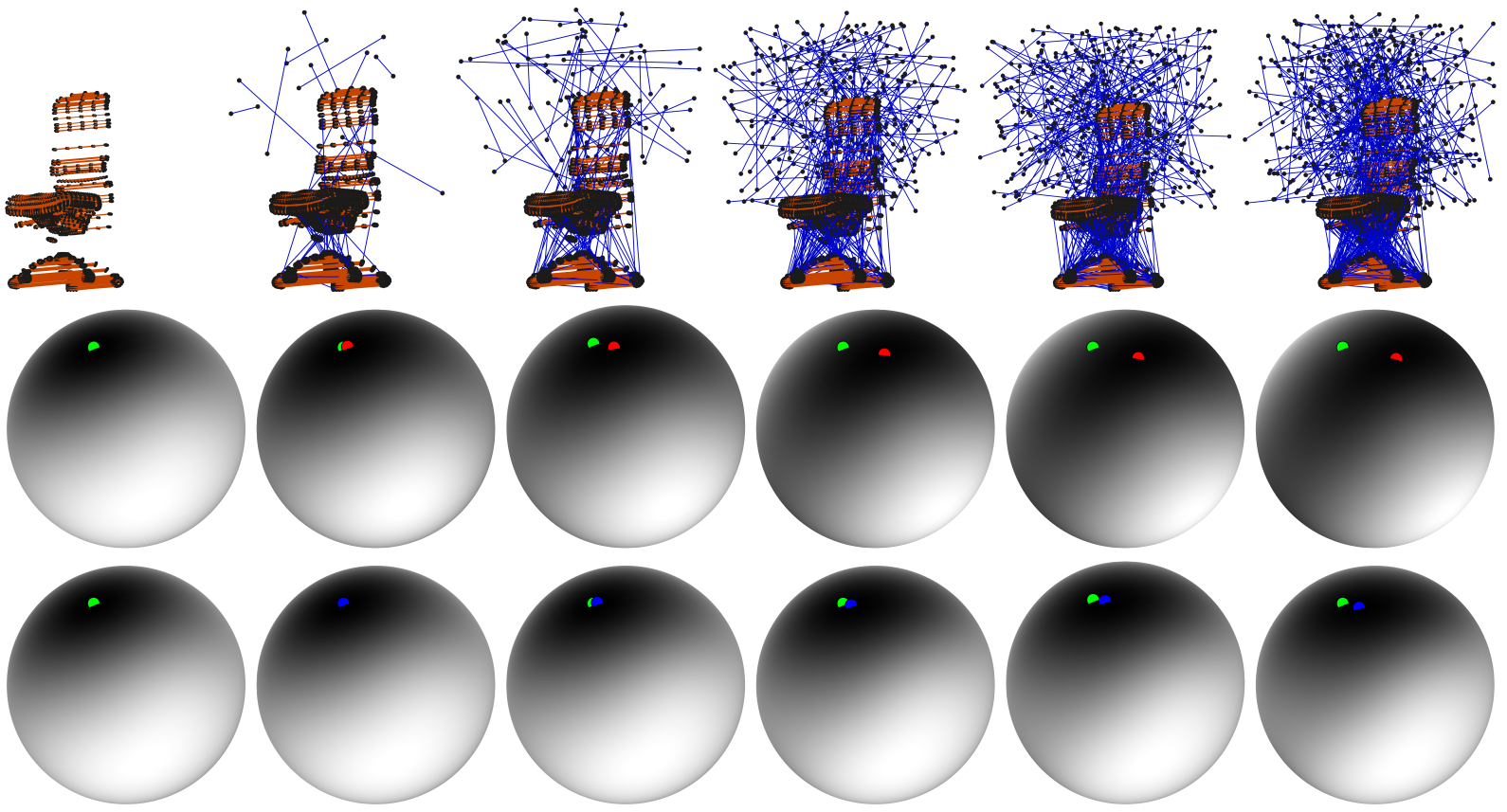,width=1\linewidth}}{}
\caption{In 1st row, we increase outlier matches (blue lines) and perturb true matches (orange lines). In 2nd row, we show the true minimum (green point) of the  function (MLE estimator) defined in Equation (\ref{eq9}). The estimated solution (red point) obtained by minimizing it quickly moves away from the true solution as we increase  outliers. This is the main limitation of existing methods (\cite{ecins2017detecting},\cite{cicconet2017finding},\cite{nagar2019detecting} ,\cite{nagar20203dsymm}) as they use L2-norm based cost function. We minimize an $L_2E$ estimator based cost function $-f$  that assigns low probabilities to many outlier and hence the estimated solution (3rd raw, blue points) remains nearby to the global solution even with many outliers. We show cost functions on $\mathbb{S}^2$.}
\label{fig:rh}
\end{figure*}
\subsection{Complete Algorithm and Analysis}
\label{sec:caa}
\begin{figure}[!h]
	\centering
	\stackunder{\epsfig{figure=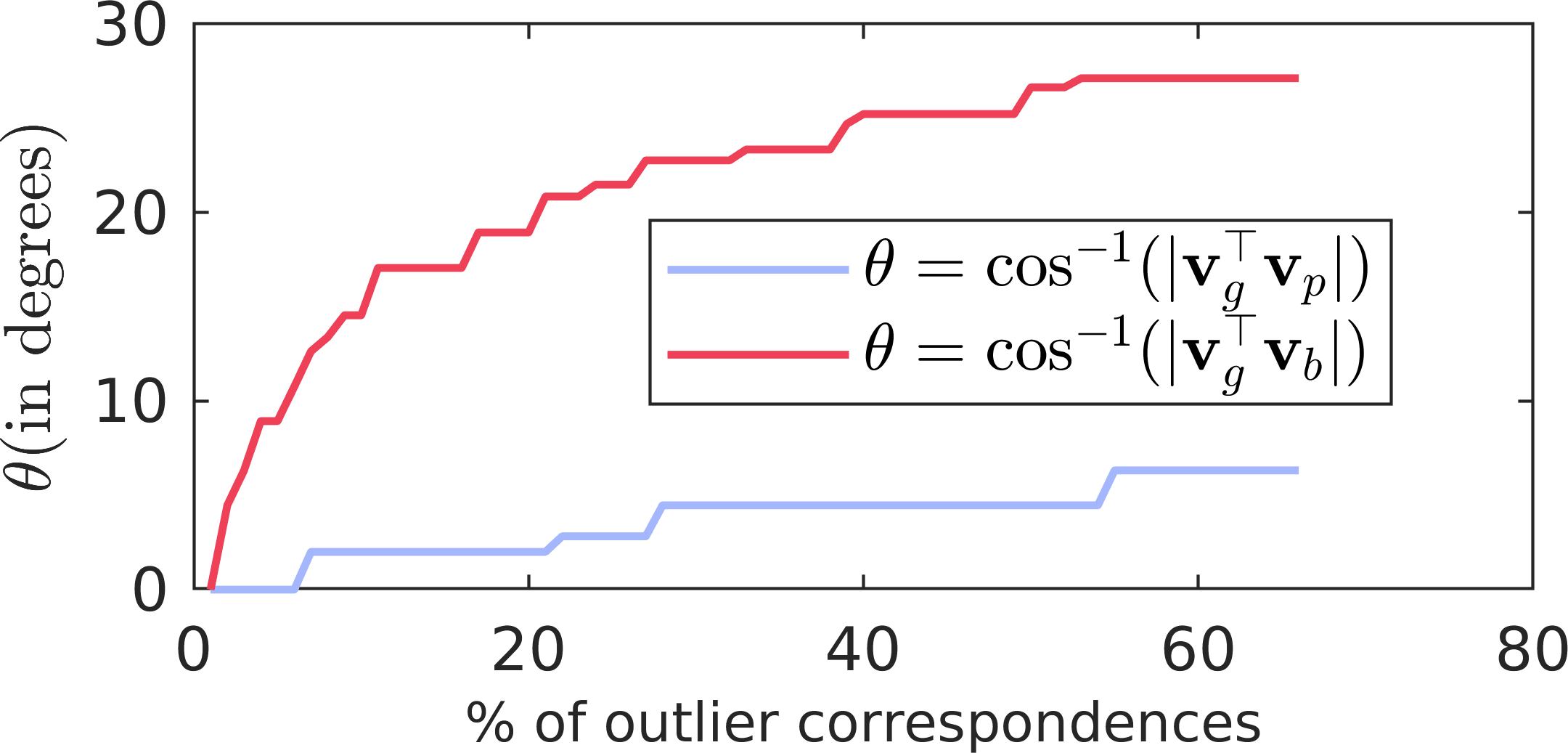,width=1\linewidth}}{}
	d\caption{The deviation angle between the estimated and ground-truth normal vectors  for the proposed approach ($\cos^{-1}(\mid\mathbf{v}_g^\top\mathbf{v}_p\mid)$) and for the baseline method ($\cos^{-1}(\mid\mathbf{v}_g^\top\mathbf{v}_b\mid)$).}
	\label{fig:dv}
\end{figure}

Now, we present the complete method for finding the normal vector to the plane of reflection symmetry given a 3D point cloud in Algorithm \ref{alg:0}. In Figure \ref{fig:rh}, we demonstrate the robustness of our approach for outliers. In the first row, we add a different number of outlier correspondences (blue color lines) and perturb the ground-truth correspondences (orange color lines) between the symmetric mirror points. In the second row, we show the actual global minimum (green color point) of the cost function defined in Equation \eqref{eq1}. We observe that if we directly optimize this cost function, the estimated solution   (red color point) quickly moves away from the global solution as we increase the number of outlier correspondences. This is the main limitation of the state-of-the-art methods (\cite{ecins2017detecting},\cite{cicconet2017finding},\cite{nagar2019detecting},\cite{nagar20203dsymm}) as they minimize the L2-norm based cost function. Whereas, we minimize the $L_2E$ cost function $-f$ defined in Equation \eqref{eq6} that can assign low probabilities to many outlier correspondences and hence the estimated solution (shown in 3rd raw by blue color) remains close to the global solution even with a significant number of outliers. Here, we have shown the cost functions on 2-Sphere $\mathbb{S}^2$. In Figure \ref{fig:dv}, we plot the deviation angle $\theta$  between the estimated normal vector and the ground-truth normal vector  for the proposed approach ($\cos^{-1}(\mid\mathbf{v}_g^\top\mathbf{v}_p\mid)$) and for the baseline method ($\cos^{-1}(\mid\mathbf{v}_g^\top\mathbf{v}_b\mid)$). Here, the baseline approach is equivalent to solving the optimization problem proposed in Equation \eqref{eq2}. We observe that the deviation for the proposed approach remains very small ($<7^\circ$) even for a large number of outlier correspondences (65\%).  
\section{Results and Evaluation}
\label{RE}

\subsection{Evaluation of Reflection Symmetry Plane}
\label{subsec:rsp}
\begin{figure}[!h]
\centering
\epsfig{figure=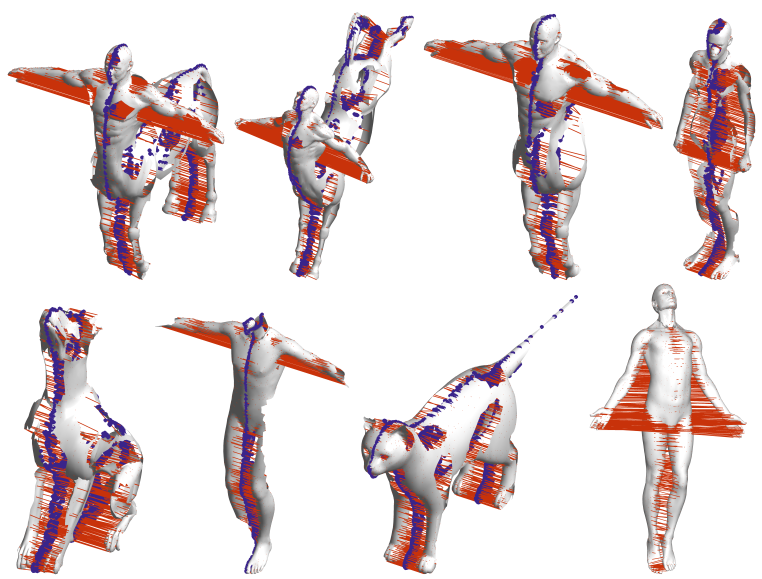,width=1\linewidth}
\caption{Results of the proposed approach on the eight models with significant missing parts from the dataset \cite{cosmo2016shrec}. }
\label{fig:res}
\end{figure}

\textbf{Benchmark Dataset and Evaluation Metric:} In order to test the performance of the proposed approach, we find the accuracy of detecting the symmetry plane. We use the benchmark dataset and the standard evaluation metric  proposed by Funk \emph{et al.} \cite{funk20172017}. Funk \emph{et al.} published a dataset of 1354  objects represented as 3D point clouds exhibiting single reflection symmetry.  We compare the accuracy of detection of plane of symmetry with the accuracies that of the state-of-the-art methods proposed by  Cicconet \emph{et al.} \cite{cicconet2017finding}, Ecins \emph{et al. } \cite{ecins2017detecting}, Speciale \emph{et al. } \cite{speciale2016symmetry},  Hruda \cite{hruda2021robust}, and   Nagar and Raman \cite{nagar20203dsymm}. We use the F-score as a performance measure metric proposed by Funk \emph{et al.} \cite{funk20172017}.  We also compare the accuracy of symmetry detection with the accuracy that of the methods proposed by Cicconet \emph{et al.} \cite{cicconet2017finding} and Nagar and Raman \cite{nagar20203dsymm} for the detection of partial and approximate symmetry detection.  We first find the precision and recall rates and then use them to find the F-score. The precision rate of detecting symmetry plane is defined as $p_r=\frac{t_p}{t_p+f_p}$,  the the recall rate  of detection of symmetry plane is defined as $r_c=\frac{t_p}{t_p+f_n}$, and the value of the  F-Score is defined as $f_s=\frac{2p_r\times r_c}{p_r+r_c}$. Here, we define the quantities $t_p$ (true positives), $f_p$ (false positives), and $f_n$ (false negatives) as follows: $t_p$ is the number of correctly detected planes of symmetry, $f_p=$ is the number of incorrectly detected planes of symmetry, and $f_n$ is the number of undetected ground-truth planes of symmetry.
In the benchmark dataset, the ground symmetry plane is defined by three points on it $\mathbf{p}_a$, $\mathbf{p}_b$, and $\mathbf{p}_c$. Let us assume that the points $\mathbf{q}_a$, $\mathbf{q}_b$, and $\mathbf{q}_b$  be any three non co-linear points on the estimated plane of symmetry.  Then, according to Funk  \emph{et al.}, the detected plane of symmetry is correct if the angle between the normals of the ground-truth  plane of symmetry   $\mathbf{g}_n=(\mathbf{p}_a-\mathbf{p}_b)\times(\mathbf{p}_a-\mathbf{p}_c)$ and the estimated  plane of symmetry $\mathbf{e}_n=(\mathbf{q}_a-\mathbf{q}_b)\times(\mathbf{q}_a-\mathbf{q}_c)$ is smaller than a predefined threshold. Also, the distance between the  center of the ground-truth plane of symmetry $\mathbf{g}_c=\frac{\mathbf{p}_a+\mathbf{p}_c}{2}$  and the center of the detected plane of reflection symmetry  $\mathbf{e}_c=\frac{\mathbf{q}_a+\mathbf{q}_c}{2}$ should be  smaller than a given threshold $t_d$ on the distance. We change the value of the threshold $t_a$ on angle criterion  in the range $\left[0,\frac{\pi}{4}\right]$ and the value of the threshold $t_d$ on the distance criterion  in the range $[0,2s]$. Here, the constant $s$ is defined to be equal to $\min\{\|\mathbf{p}_a-\mathbf{p}_b\|_2,\|\mathbf{p}_a-\mathbf{p}_c\|_2,\|\mathbf{q}_a-\mathbf{q}_b\|_2,\|\mathbf{q}_a-\mathbf{q}_c\|_2\}$. Since this dataset contains perfectly symmetric objects with no outliers, we define the center of the plane of symmetry as the mean center of the input point cloud. Hence, the distance threshold criterion is true for all the approaches.\begin{figure}[!h]
\centering
\stackunder{\epsfig{figure=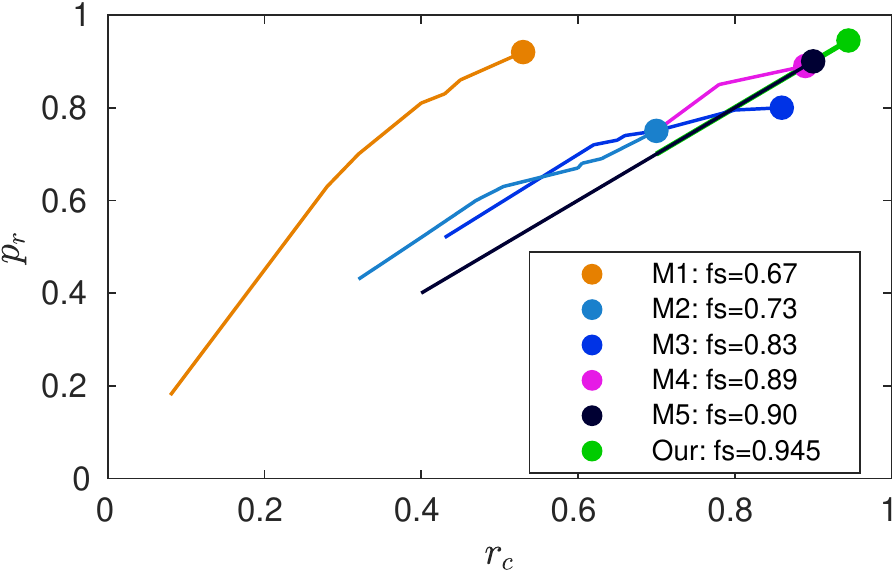,width=1
		\linewidth}}{}
\caption{Precision $(p_r)$ vs Recall $(r_c)$ curve for the methods proposed by Cicconet \emph{et al.} \cite{cicconet2017finding} (M1), Ecins \emph{et al. } \cite{ecins2017detecting} (M2), Speciale \emph{et al. } \cite{speciale2016symmetry} (M3), Hruda \cite{hruda2021robust} (M4), Nagar and Raman \cite{nagar20203dsymm} (M5), and our method on the benchmark dataset \cite{funk20172017}.}
\label{fig:eval}
\end{figure}
In Figure \ref{fig:eval}, we plot the recall vs precision curves for  the algorithms proposed by  Cicconet \emph{et al.} \cite{cicconet2017finding}, Ecins \emph{et al. } \cite{ecins2017detecting}, Speciale \emph{et al. } \cite{speciale2016symmetry},  Hruda \cite{hruda2021robust}, and   Nagar and Raman \cite{nagar20203dsymm}. We observe that the value of the  F-score (0.93) for the proposed approach is the highest among all methods. The benchmark dataset \cite{funk20172017} contains complete objects. Therefore the performance of the method proposed in \cite{nagar20203dsymm} is comparable to the performance of the proposed approach.
In Figure \ref{fig:res}, we plot the estimated normal vectors to symmetry planes, a few sampled points (red color points) on the detected plane, and correspondences between mirror symmetric points (blue color lines) using the proposed approach for a few 3D point clouds from the dataset \cite{cosmo2016shrec}.  We observe that our method can detect symmetry of  objects using their partial scans in the presence of outlier points. We have also used our algorithm on dynamic model (last model of second row of Figure \ref{fig:res}). We observe that the proposed approach is able to detect the symmetry in the extrinsically symmetric parts (hands for the first model and legs for the second model). There are some mirror symmetric correspondences (as HKS is invariant to isometric deformations) detected in the non-symmetric region, but they are not accurate enough. Actually, the dynamic models exhibit intrinsic symmetry, and there is a separate category of algorithms for solving this problem, e.g. \cite{nagar2018fast,ovsjanikov2008global,kim2010mobius}. 
\subsection{Evaluation of Partial Symmetry Detection} 
\begin{figure}[!h]
\centering
\stackunder{\epsfig{figure=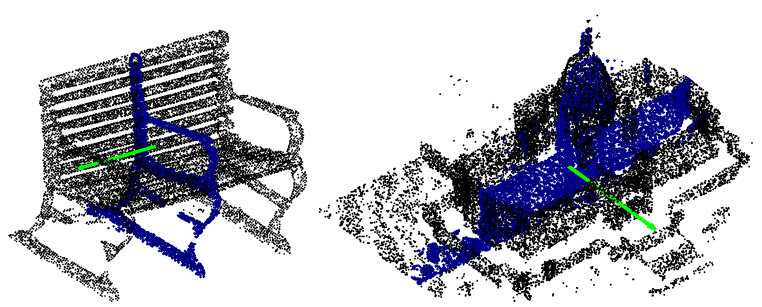,width=1\linewidth}}{}
\caption{Results of the proposed approach on the two models from the partial and real scan dataset \cite{funk20172017}.}
\label{fig:res1}
\end{figure}

In order to measure the performance of our approach and compare it with the performances of the other state-of-the-art methods  \cite{nagar20203dsymm} and \cite{hruda2021robust} on 3D objects with missing parts, we follow the same evaluation process proposed in Section \ref{subsec:rsp} and the dataset proposed in \cite{funk20172017}. This dataset has partial scans of 20 real-world objects along with their ground-truth plane of symmetry.  The average value of the F-Score for the methods in \cite{nagar2019detecting}, \cite{hruda2021robust}, and for  proposed algorithm are 0.90, 0.93, and 0.96, respectively. We obtain higher F-score for partial scan than that for the complete scans since the partial scan dataset contains only 20 models whereas the dataset for complete scans contains 1354 models. In Figure \ref{fig:res1}, we show detected planes for two model from this dataset.

\begin{table*}[!h]
	\caption{The mean ground truth error (GTE) and the mean symmetry distance error (SDE) of for the different methods on the 1000 models of different categories of the ShapeNet \cite{chang2015shapenet} dataset.}
	\begin{center}{\scriptsize
			\begin{tabular}{cccccccccc}
				Error& PCA  &  Kazhdan   & Martinet   & Mitra   & PRST  & Korman  & PRS-Net   &RANSAC& Ours  \\
				&    &  \cite{kazhdan2002reflective} &  \cite{martinet2006accurate}  &  \cite{mitra2006partial} & \cite{podolak2006planar}  &  \cite{korman2015probably} & \cite{gao2020prs}  && New  \\
				\hline
				GTE($\times10^{-2}$)&2.41    &0.17  &13.6  &52.1  &4.42    &19.2  & \textbf{0.11} &1.53& 0.13  \\\hline
				SDE($\times10^{-4}$)&3.32    &0.897  &3.95  &14.2  &1.78    &1.75 & \textbf{0.86} &2.31&0.89  \\
				\hline
		\end{tabular}}
	\end{center}
	\label{gtesde}
\end{table*}
\subsection{Symmetry Distance Error and Ground-Truth Error}
We compare the performance of the proposed approach with a recent state-of-the-art learning based algorithm PRSNet \cite{gao2020prs} and classical algorithms \cite{kazhdan2002reflective},  \cite{martinet2006accurate},  \cite{mitra2006partial},  \cite{podolak2006planar},   and \cite{korman2015probably}  on the ShapeNet dataset \cite{chang2015shapenet}. We follow the evaluation metrics symmetry distance error (SDE) and ground-truth error (GTE) as proposed in \cite{gao2020prs}. The SDE is defined as the distance between the original point cloud and the reflected point cloud using the detected symmetry plane. The GTE is defined as the euclidean distance between the distance between the  ground truth symmetry plane and detected symmetry plane. In Table \ref{gtesde}, we present the SDE and the GTE for these approaches. We observe that the proposed approach has better performance than the classical approaches and comparable performance with that of the PRSNet algorithm (a supervised learning based approach) and Kazhdan et al. \cite{kazhdan2002reflective}.  	
\subsection{Effect of Outliers and Missing Parts} 
\begin{figure}[!h]
\centering
\stackunder{\epsfig{figure=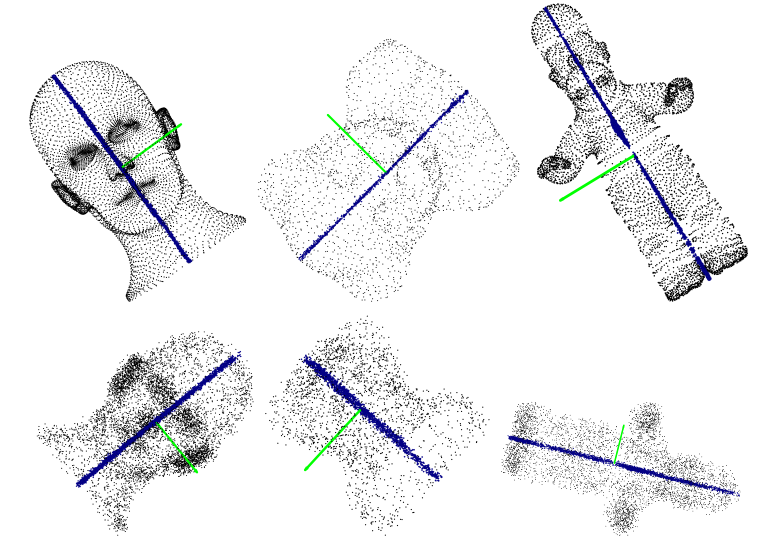,width=1\linewidth}}{}
\caption{The first row represents the detected symmetries in clean models. The second row represents the detected symmetries in noisy models (models are from the work \cite{hruda2021robust}).}
\label{fig:noisy}
\end{figure}

\begin{figure}[!h]
\centering
\stackunder{\epsfig{figure=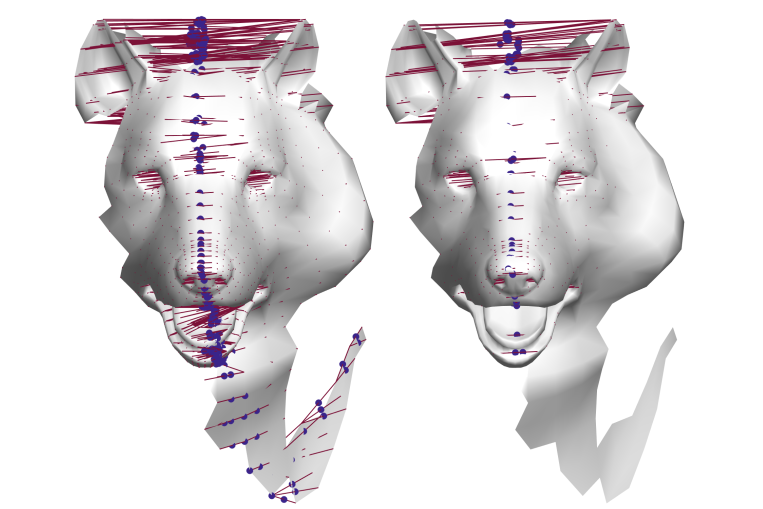,width=1\linewidth, angle =0}}{}
\caption{Left: Result of the approach proposed in \cite{nagar20203dsymm} on a partial model. Right: Result of the proposed approach on the same partial model.}
\label{fig:intro1}
\end{figure}

In order to measure the performance  in presence of a significant amount of outliers, we use the  strategy proposed in \cite{nagar20203dsymm}. We again consider the 3D objects provided in dataset \cite{funk20172017} and introduce random points  into the input 3D point cloud to get a modified  noisy point cloud $\mathcal{P}^\text{n}=\mathcal{P}\cup\mathcal{P}_\text{r}$. Here, the point cloud  $\mathcal{P}_\text{r}$ is a set of outliers. We choose different number of  noise points such that  $\mid\mathcal{P}_\text{r}\mid=\frac{\alpha}{100}\mid\mathcal{P}\mid, \;\alpha\in\{0,20,\ldots,120\}$.  
We find the value of the F-score for the method proposed in  \cite{nagar20203dsymm} and the proposed algorithm for every value of the new point cloud $\mathcal{P}^\text{n}$. We use the dataset \cite{funk20172017} for evaluation. In Table \ref{tab:1} present the obtained F-score values for different values of $\alpha$. We observe that even for the case where half  of the points are outliers, the F-score for the proposed approach remains around 0.88. 
\begin{table}[!h]
\caption{The value of F-score for the proposed approach and  \cite{nagar20203dsymm} as we vary the value of $\alpha$ in $\{0,20,\ldots,100\}$.}
\begin{center}
	\begin{tabular}{ccccccc}
		\hline
		$\alpha\rightarrow$& 0 & 20 & 40 & 60 &80  &100    \\ 
		\hline 
		\cite{nagar20203dsymm} &0.90&0.87  &0.86  &0.86  & 0.85 & 0.85    \\ 
		\hline 
		Proposed&0.95  &0.94  &0.92  &0.91  &0.91  &0.89        \\ 
		\hline 
	\end{tabular} 
\end{center}
\label{tab:1}
\end{table}
To measure the performance of our method and compare it with that of the state-of-the-art methods for partial symmetry, we remove a set $\mathcal{P}_{m}$ of connected points  from the input 3D point cloud $\mathcal{P}$ such that $\mid\mathcal{P}_m\mid=\gamma\mid\mathcal{P}\mid,\;\gamma\in\{0,0.15,0.20,0.28\}$ and find the value of F-score for the method proposed in \cite{nagar20203dsymm} and the proposed algorithm. We use the dataset \cite{funk20172017} for evaluation. In Table \ref{tab:2}, we  present the obtained F-score values for different values of $\gamma$. We note that even for $\gamma=0.28$,  F-score  for the proposed approach remains 0.89.  In Figure \ref{fig:noisy}, we show detected planes for a few models from the dataset \cite{hruda2021robust} to show that the proposed approach is robust to noisy points.  We present a  qualitative result compared with \cite{nagar20203dsymm} in Figure \ref{fig:intro1}. The main reason for the failure of the method \cite{nagar20203dsymm} is that the symmetry plane detection algorithm fails to converge to the final solution if the angle between the initialized normal vector and the ground-truth normal vector is more than around $20^\circ$. We observe that there are many outlier matches to a part (neck and chest)  for which the mirror part does not exist.  In the presence of many outliers and missing parts, this initialize algorithm fails to properly initialize the normal vector. Whereas, the proposed estimator is significantly robust (angle error $<7\%$) to many outlier correspondences (65\%) and produces good matches at the convergence.
\begin{table}[!h]
\caption{The value of F-score for the proposed approach and \cite{nagar20203dsymm} as we vary the value of $\gamma$ in the range $\{0,0.15,0.20,0.28\}$.}
\begin{center}
	\begin{tabular}{ccccc}
		\hline 
		$\gamma\rightarrow$& 0 & 0.15 & 0.20 & 0.28   \\ 
		\hline 
		\cite{nagar20203dsymm} &0.90&0.82  &0.75  &0.67    \\ 
		\hline 
		Proposed &0.95  &0.93  &0.92  &0.90   \\ 
		\hline 
	\end{tabular} 
\end{center}
\label{tab:2}
\end{table}
\subsection{Time Complexity:} The BFGS method takes $\mathcal{O}(n)$ \cite{absil2009optimization}. The descriptor finding step takes $\mathcal{O}(nh^2)$,   ($\mathcal{O}(h^2)$ for finding eigenvalues of $\mathbf{D}$). Here, $h$ is the average number of neighboring points. For our experiments $h\approx100$. Therefore, the overall approximate time complexity of our approach is $\mathcal{O}(n)+\mathcal{O}(nh^2)$. The time complexity of the method proposed in \cite{nagar2019detecting} is $\mathcal{O}(n^{3.5})$. For a model with 5k points, $k=20$ and $r=0.1$, our algorithm takes around 1.58 seconds on a Linux OS with an Intel-i7 processor.  We report the time comparison in Table \ref{tab:3}. We observe that the computation time for the method proposed in \cite{nagar20203dsymm} increases drastically ($\mathcal{O}(n\log n)$) for large point clouds as it solves a nearest-neighbor search problem for updating mirror correspondences in each iteration. Our algorithm finds a robust estimator for the symmetry plane even from noisy mirror correspondences and does not require solving the nearest-neighbor problem. We need to run Manifold-BFGS solver to find $\mathbf{v}$ which has only two variables. Hence, our computational complexity is almost linear in point cloud size. \
\begin{table}[!h]
\caption{Computation time comparison with \cite{nagar20203dsymm} on the dataset \cite{funk20172017}.}
\begin{center}
	\begin{tabular}{cccccc}
		\hline 
		$\#$Vertices& 5k &10k  &50k&100k&300k\\ 
		\hline 
		\cite{nagar20203dsymm}&  0.38s&0.48s & 6.12s&16.03s& 61.01s\\ 
		\hline 
		Proposed&2.15s & 3.67s& 5.23s & 8.13s&12.92s \\ 
		\hline 
	\end{tabular}
\end{center}
\label{tab:3}
\end{table}
We have also compared the computation time for the proposed approach with that of the methods  proposed in  Korman \emph{et al.}   \cite{korman2015probably}, Mitra \emph{et al.} \cite{mitra2006partial}, and Gao \emph{et al.}   \cite{gao2020prs}. We have used i7 processor for comparing the computation time. We have used 3D models having around 1042  vertices. In Table \ref{timcop}, we present the computation for these approaches. We observe that the proposed approach has the second best computation time. It is the best among non-learning based approaches. The method preposed in \cite{gao2020prs} achieves the best test computation time which is a learning based approached.
\begin{table}
\caption{Computation time (in sec) for  Korman \emph{et al.}   \cite{korman2015probably}, Mitra \emph{et al.} \cite{mitra2006partial}, Gao \emph{et al.}   \cite{gao2020prs}, and proposed approach.}
\begin{center}
	\begin{tabular}{ccccc}
		\hline
		Method  & \cite{korman2015probably} & \cite{mitra2006partial}   &\cite{gao2020prs} & Proposed  \\
		\hline
		Time (s)& 0.97 & 0.42 &0.0018& 0.21\\
		\hline
	\end{tabular}
\end{center}
\label{timcop}
\end{table}
\subsection{Missing Part Completion using the  Detected Symmetry}
\begin{figure}[!h]
\centering
\stackunder{\epsfig{figure=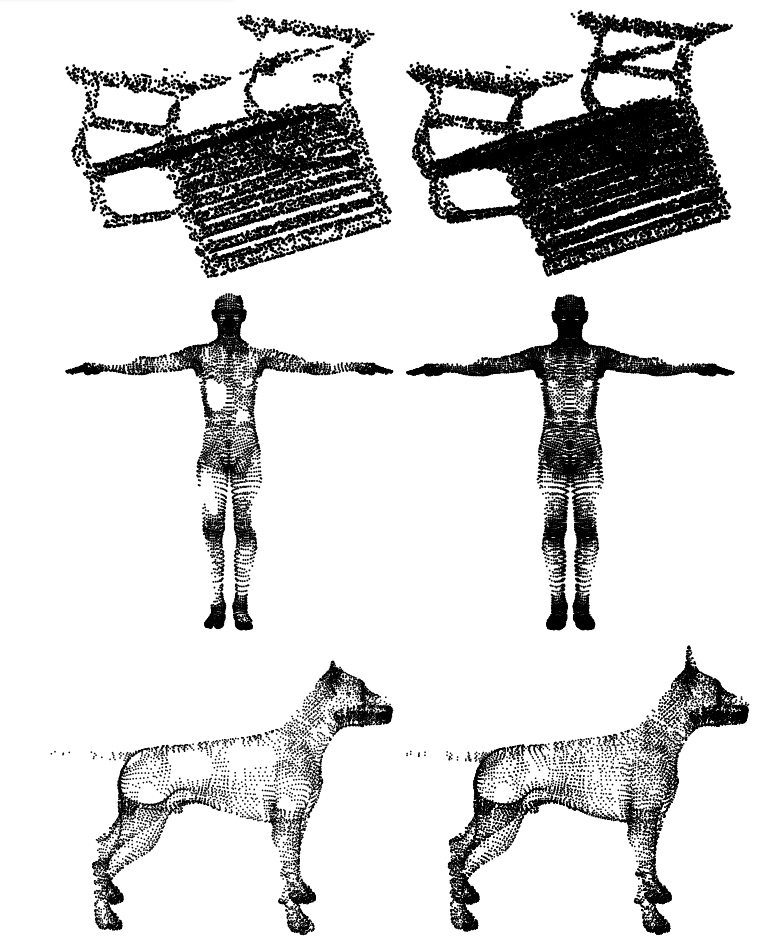,width=1\linewidth}}{}
\caption{Object completion using the detected reflection symmetry.First Column: Incomplete Objects. Second Column: Symmetry Guided Completion.}
\label{fcomp1}
\end{figure}

In order to show the practical application of the proposed approach, we use the detected symmetry of partial symmetric objects to reconstruct the missing parts.  Let $\mathcal{P}=\{\mathbf{x}_i\}_{i=1}^n$ be an incomplete point cloud exhibiting reflection. Let $\mathbf{v}$ be the estimated normal vector to the symmetry plane of the object represented by $\mathcal{P}$. Then, we reconstruct the complete object as $\mathcal{P}\cup\mathcal{Q}$ where $\mathcal{Q}=\{(\mathbf{I}-2\mathbf{vv^\top})\mathbf{x}_i+2\omega\mathbf{v}\}_{i=1}^n$. In Figure \ref{fcomp1}, we show a few partial objects and the completed objects using this approach.

\section{Conclusion}
\label{sec:CFI}
In this work, we have proposed a fast and robust algorithm for detecting symmetry of 3D objects exhibiting single reflection symmetry and represented by partial and noisy points clouds. We adapted a statistical estimation technique for finding the plane of reflection symmetry. For this purpose, we first found a 3D point descriptor for each point that is invariant to reflection symmetry transformation. Then, we used an approximate nearest neighbor matching technique for finding a set of candidate correspondences between mirror reflective points. We used this  set of putative correspondences and a statistical estimator to estimate the reflection symmetry plane that is robust to a significant number of outliers and missing parts. The proposed approach achieved comparable mean ground-truth error and 4.5\% increment in the F-score as compared to the state-of-the-art approaches on the benchmark dataset.

\textbf{Limitations and Future Works:} The proposed work only detects the global intrinsic reflection symmetry of a given 3D model represented by a point cloud. However, many objects may reflect local symmetry along with global symmetry, e.g. a building. In future work, we would like to extend our framework for detecting symmetries objects exhibiting multiple reflection symmetries. Furthermore, the proposed approach detects extrinsic reflection symmetry in a point cloud. We would further like to extend the proposed approach for detecting intrinsic symmetries of non-rigid objects. \\
\textbf{Acknowledgments:} This work was funded by the SERB, Government of India with through
the Start-up Research Grant (SRG) scheme.



\end{document}